\pdfoutput=1

\documentclass[11pt]{article}

\usepackage[final]{acl}

\usepackage{times}
\usepackage{latexsym}

\usepackage[T1]{fontenc}

\usepackage[utf8]{inputenc}

\usepackage{microtype}

\usepackage{inconsolata}

\usepackage{graphicx}

\usepackage{lipsum}
\usepackage{colortbl}

\usepackage{tabularx}
\usepackage{booktabs}
\usepackage{adjustbox}
\usepackage{multirow}

\usepackage{enumitem}
\usepackage{float}

\usepackage{amsmath}

\definecolor{uzhblue}{RGB}{0,40,165}
\usepackage{tcolorbox}

\usepackage{listings}

\usepackage{xcolor}

\lstdefinelanguage{json}{
  basicstyle=\ttfamily,
  numbers=left,
  numberstyle=\tiny\color{gray},
  stepnumber=1,
  numbersep=5pt,
  stringstyle=\color{black},
  showstringspaces=false,
  morestring=[b]",
  literate=
   *{0}{{{\color{black}0}}}{1}
    {1}{{{\color{black}1}}}{1}
    {2}{{{\color{black}2}}}{1}
    {3}{{{\color{black}3}}}{1}
    {4}{{{\color{black}4}}}{1}
    {5}{{{\color{black}5}}}{1}
    {6}{{{\color{black}6}}}{1}
    {7}{{{\color{black}7}}}{1}
    {8}{{{\color{black}8}}}{1}
    {9}{{{\color{black}9}}}{1}
}

\usepackage{caption}
\usepackage{url}

\title{The \textit{Mediomatix} Corpus: Parallel Data for\\ Romansh Language Varieties via Comparable Schoolbooks}

\author{
  \textbf{Zachary Hopton\textsuperscript{1}} \quad
  \textbf{Jannis Vamvas\textsuperscript{1}} \quad
  \textbf{Andrin Büchler\textsuperscript{2}} \\
  \textbf{Anna Rutkiewicz\textsuperscript{1}} \quad
  \textbf{Rico Cathomas\textsuperscript{2}} \quad
  \textbf{Rico Sennrich\textsuperscript{1}}\\[0.2em]
  \textsuperscript{1}University of Zurich \quad
  \textsuperscript{2}University of Teacher Education of the Grisons
\\
  \small{
    \href{mailto:zacharywilliam.hopton@uzh.ch}{zacharywilliam.hopton@uzh.ch},  \href{mailto:vamvas@cl.uzh.ch}{vamvas@cl.uzh.ch}
  }
}

\begin{document}
\maketitle
\begin{abstract}
  The five idioms (i.e., varieties) of the Romansh language are largely standardized and are taught in the schools of the respective communities in Switzerland.
  In this paper, we present the first parallel corpus of Romansh idioms.
    The corpus is based on 291 schoolbook volumes, which are comparable in content for the five idioms.
We use automatic alignment methods to extract 207k multi-parallel segments from the books, with more than 2M tokens in total.
  A small-scale human evaluation confirms that the segments are highly parallel, making the dataset suitable for NLP applications such as machine translation between Romansh idioms.
  We release the parallel and unaligned versions of the dataset under a CC-BY-NC-SA license\footnote{Parallel Corpus: \url{https://huggingface.co/datasets/ZurichNLP/mediomatix}; Unaligned Corpus: \url{https://huggingface.co/datasets/ZurichNLP/mediomatix-raw}}
  and demonstrate its utility for machine translation by training and evaluating an LLM and a supervised multilingual MT model on the dataset.
\end{abstract}

\section{Introduction}
Romansh, a Romance language spoken in southeast Switzerland, has five main regional varieties, which are commonly called \textit{idioms}.
Even though the variability of Romansh idioms takes a central role in the Romansh language community, including in school education, there has been no parallel corpus to date, limiting the prospect of machine translation~(MT) between Romansh idioms.

In this paper, we present the first (multi-)parallel corpus for the five Romansh idioms, which we extracted from 291 comparable schoolbooks of the \textit{Mediomatix} series for teaching Romansh as a first language.
Figure~\ref{fig:example-segments} presents an example of a short parallel segment across the five idioms, illustrating the variance between the idioms in terms of lexicon, syntax, and orthography.
In total, our multi-parallel corpus contains 207k aligned segments of varying lengths, spanning more than 2 million tokens.
We release it to the research community with permission from the schoolbook editors.

We automatically align the segments by using VecAlign~\cite{thompson-koehn-2019-vecalign}, a standard approach for embedding-based parallel sentence alignment.
To enable a high-precision multi-parallel alignment across the five idioms, we use an approach that we call \textit{pivot consensus alignment}.
We validate the alignment (1) by reporting accuracy on a validation set, and (2) by performing a small-scale evaluation through a native Romansh speaker, which showed that 471 out of 472 evaluated segments were aligned correctly.

We further demonstrate that our corpus can be successfully used for fine-tuning NLP systems on the task of MT from one idiom into another, which, to our knowledge, is the first such attempt for Romansh idioms.
Code to reproduce our experiments is available.\footnote{\url{https://github.com/ZurichNLP/mediomatix-code}}

\begin{figure}[t]
\centering
\begin{tabular}{>{\columncolor[HTML]{DBEDAD}}l>{\columncolor[HTML]{ECF6D6}}l}
   & \\[-0.9em]
  \textbf{Sursilvan:} & \textit{Speronza ei quei ver.} \\[0.1em]
  \textbf{Sutsilvan:} & \textit{Sprànza e quegl ver.} \\[0.1em]
\textbf{Surmiran:} & \textit{Speranza è chegl veir!} \\[0.1em]
\textbf{Puter:} & \textit{Spraunza es que vaira!} \\[0.1em]
\textbf{Vallader:} & \textit{Spranza es quai vaira!} \\[0.2em]
[English] & \textit{Hopefully that's true!} \\[0.3em]
\end{tabular}
\caption{Example of a parallel segment in the five Romansh idioms.}
\label{fig:example-segments}
\end{figure}

\begin{table*}
\centering
\begin{tabularx}{\textwidth}{@{}Xr@{\hspace{4em}}rr@{\hspace{4em}}rr@{}}
\toprule
\textbf{Idiom} & \textbf{Book volumes} & \multicolumn{2}{l}{\textbf{Segments}} & \multicolumn{2}{l}{\textbf{Tokens}} \\
& (workbook + commentary) & Overall & Aligned      & Overall & Aligned \\ \midrule
Sursilvan & 67 & 104,481 & 49,872 & 955,606 & 513,036 \\
Sutsilvan & 67 & 104,825 & 49,137 & 989,282 & 513,850 \\
Surmiran & 27 & 51,723 & 14,301 & 463,694 & 145,982 \\
Puter & 65 & 106,348 & 47,185 & 993,705 & 494,850 \\
Vallader & 65 & 113,754 & 47,397 & 1,058,343 & 495,404 \\
\midrule
\textbf{Total} & \textbf{291} & \textbf{481,131} & \textbf{207,892} & \textbf{4,460,630} & \textbf{2,163,122} \\ \bottomrule
\end{tabularx}
\caption{Dataset statistics for the \textit{Mediomatix} corpus.}
\label{tab:stats}
\end{table*}

\section{Background}
\label{sec:background}
\subsection{Language Situation}
Alongside German, French, and Italian, Romansh is one of Switzerland's four national languages (ISO 639-1: \texttt{rm}; ISO 639-2/3: \texttt{roh}). It has an estimated number of 60,000 speakers~\cite{mueller2021_schweiz} and enjoys legally protected minority status~\citep{etter2018widerspruche}.
Romansh's present-day situation is strongly influenced by its unique sociolinguistic context~\citep{grunert2024ratoromanisch}. The ``traditional'' speaking area in the canton of Grisons comprises five core territories which are divided geographically but also linguistically.

\paragraph{Romansh Idioms}
Each of these territories has its ``own'' Romansh idiom (i.e., regional standard variety). These idioms are \textbf{Sursilvan} (estimated 55\% of Romansh speakers) in western Grisons, \textbf{Sutsilvan} (estimated 3\% of Romansh speakers) and \textbf{Surmiran} (estimated 10\% of Romansh speakers) in central Grisons, \textbf{Puter} (estimated 12\% of Romansh speakers) in the upper Engadine valley, and \textbf{Vallader} (estimated 20\% of speakers) in the lower Engadine valley \citep{furer2005aktuelle}. Each of the Romansh idioms are standardized, written forms of the language that have emerged over the last 400 years~\cite{caviezel1993geschichte,liver2010ratoromanisch}
and which have their own codices (e.g., \citealp[]{spescha89grammatica}) and literary traditions. There are therefore substantial differences not only in orthography but also other linguistic levels such as vocabulary (e.g., \textit{glimaglia} in Sursilvan vs. \textit{lindorna} in Vallader for ‘snail’) or morphosyntax (e.g., analytic future tense using the auxiliary verb \textit{to come} in Sursilvan and Sutsilvan vs. synthetic future tense in the other idioms). As a result, mutual intelligibility is sometimes challenging for speakers of differing idioms.

\paragraph{Contemporary Role of Romansh Idioms} Evolution of five differing regional standard varieties was fostered by factors such as processes of demarcation to avoid intermixing of idioms~\cite{arquint1982_standardization} or strong regional attachment and closeness to the vernacular~\citep{diekman1991probleme}.
In the 1980s, linguists and policy makers created and implemented a supraregional standard variety \citep{cathomas2024weg,coray2010rumantsch,schmid89richtlinien}. This variety, known as \textit{Rumantsch Grischun}, entails elements from all idioms~\citep{gross1999rumantsch}. Nowadays, Rumantsch Grischun is used partly in media, in higher education, and in federal and cantonal administration~\citep{grunert2024ratoromanisch} but only in a very limited number of schools as language of instruction. Consequently, the broader population is not literate in Rumantsch Grischun but in one of the five idioms.

\subsection{NLP for Romansh}
While Romansh idioms have been studied in the context of commercial speech technology\footnote{\url{https://recapp.ch/}}, previous work on the processing of Romansh text has focused on Rumantsch Grischun, the supraregional standard form of the language \citep[e.g.,][]{muller-etal-2020-domain, dolev-2023-mbert,vamvas-etal-2023-swissbert}.
With the \textit{Mediomatix} corpus, we hope to provide a basis for multilingual NLP research on written text in the five regional idioms of Romansh.

\subsection{Teaching Materials as a Source of Parallel Text for Romansh Idioms}

The five Romansh idioms are used as school languages in their respective areas,
alongside German as an equal or dominant language of instruction.
Additionally, Romansh is explicitly taught as a subject~\citep{cathomas2005schule,gross2017romansh}.
For the latter case, a series of teaching materials is being developed for grades 2 to 9 (i.e., age 8 to 16) at the University of Teacher Education of the Grisons, called \textit{Mediomatix}.\footnote{\url{https://mediomatix.ch/}}


Once complete, \textit{Mediomatix} will comprise 325 components for teaching and learning: 160 workbooks (32 per idiom), 160 commentaries (32 per idiom) and five grammar books. In total, this sums to 16,000 pages containing around 80,000 structural elements (texts, tasks, exercises, images, instructions, charts, links, etc.). This large compilation of textual data in different idioms is representative of Romansh school or academic language (i.e., more formal than everyday speech). Because several language experts are involved in writing and proofreading, the material is highly compliant with the language norms found in the idioms’ codices.

The completion of \textit{Mediomatix} is scheduled for 2029, but more than 150 volumes are already available~(Appendix~\ref{sec:book-list}), which we believe to be a suitable basis for compiling a well-controlled, near-parallel corpus of Romansh idioms.

\section{Dataset Creation}

We create the corpus with standard NLP tools, dividing the process into two steps: (1) text extraction and segmentation; (2) embedding-based alignment of the segments with Vecalign~\citep{thompson-koehn-2019-vecalign}, using a \textit{pivot consensus alignment} strategy to create a consistent multi-parallel alignment across all five idioms.

\subsection{Extraction of Text Segments}
\label{sec:extraction}

We extract the text from the content management system used for editing the schoolbooks, so OCR is not necessary.
Among the extracted content, we retain the original content with all HTML markup, as well as the plain text extracted from the HTML.
For splitting the text into segments we follow the HTML markup, treating every paragraph, list item, etc.\ as a separate segment, and do not perform further sentence splitting.


\subsection{Multi-parallel Alignment}

Our goal is to create a multi-parallel corpus, where each segment in one idiom is aligned with its corresponding segment in the other four idioms, such as in Figure~\ref{fig:example-segments}.
To reduce complexity, we break down multi-parallel alignment into a series of pairwise alignments, which we then aggregate into a single multi-parallel alignment.

\paragraph{Bilingual Alignments}
We manually align chapters within each schoolbook volume based on their titles, and then perform automatic bilingual segment alignment within each of the aligned chapters.
We do so using the Vecalign algorithm, as previous work shows its effectiveness in low-resource settings that are near-monotonic~\cite{signoroni-rychly-2023-evaluating}.
Using the ``maximum alignment size'' hyperparameter of Vecalign, we limit the algorithm to outputting just 1--1 alignments and deletions (i.e., 1--0 and 0--1) to increase the precision of the alignment.

\paragraph{Choice of Embedding Model} We manually construct a multi-parallel validation set of~150 rows to help us choose a suitable strategy for creating segment embeddings. The manual alignment is done by an author using the PDF schoolbooks and a multilingual, multi-idiom Romansh dictionary\footnote{\url{https://www.mypledari.ch/}} as references. This allowed us to use formatting cues and English translations to cleanly align the validation set. Using this validation set, we experiment with several embedding models, both open-source and commercial (see Appendix~\ref{sec:val-exp} for experiment details and Appendix \ref{ap:val_set_stats} for details about the validation set).
Based on the validation results, we decide to use Cohere's \texttt{embed-v4.0}\footnote{\url{https://docs.cohere.com/docs/cohere-embed}} to align the full corpus.

\paragraph{Pivot Consensus Alignment}
\label{sec:pivot}
To combine the bilingual alignments into a single, multi-parallel alignment, we use a \textit{pivot} idiom (Figure \ref{fig:pivot_fig}).
Let $i$ and $j$ be two idioms, and $p$ a pivot idiom.
The \textit{pivot alignment} is the set of segments in $i$ and $j$ that are aligned via $p$: \\
$A_{ij}^{(p)} = \left\{ (s_i, s_j) | (s_i, s_p) \in A_{ip} \land (s_p, s_j) \in A_{pj} \right\}$. We include pivot-side deletions by adding unmatched segments from $A_{ip}$ and $A_{pj}$ to $A_{ij}^{(p)}$ with null counterparts. The multi-parallel alignment is then the union over all $i, j$ pairs: $\mathcal{A}^{(p)} = \bigcup_{\, i, j \, \in \, \text{idioms}} A_{ij}^{(p)}$. See Appendix \ref{ap:mp_fig} for an example visualization of multi-parallel alignment via a pivot idiom. 

Experiments on the validation set~(Table~\ref{tab:valid_f1}) indicate that pivot alignments have high recall, e.g., 99.2 when using Sursilvan as the pivot language.
However, we want to give a higher weight to precision than to recall in order to minimize the number of misaligned segments in the final corpus.
We find that precision can be increased by aggregating the five pivot alignments into a single multi-parallel alignment, using a consensus-based approach.

Specifically, we calculate the pivot consensus alignment as the intersection of the five pivot alignments:
$\mathcal{A}^{\text{consensus}}=\bigcap_{p\in \text{idioms}}\mathcal{A}^{p}$.

\paragraph{Length Heuristic}
Similar to \citet{ng-etal-2019-facebook}, we filter out segments with mismatching lengths. Segments that are 1.5 times longer or 0.67 times shorter than the average length in a row are removed from that row.

\begin{table}
    \centering
    \begin{tabularx}{\linewidth}{@{}Xrrr@{}}
        \toprule
        \textbf{Approach} & \textbf{Prec.} & \textbf{Rec.} & \textbf{F1} \\
        \midrule
        Pivoting via Sursilvan only & 94.8 & 99.2 & 96.9 \\
        Consensus across all pivots & 97.2 & 94.1 & 95.4 \\
        \bottomrule
    \end{tabularx}
    \caption{Taking the consensus across all five possible pivot languages increases precision on the validation set, compared to using a single, arbitrary pivot language.
    }
    \label{tab:valid_f1}
\end{table}

\begin{table*}
\centering
\begin{tabularx}{\textwidth}{@{}l*{5}{>{\centering\arraybackslash}X}*{5}{>{\centering\arraybackslash}X}>{\centering\arraybackslash}X@{}}
\toprule
\textbf{System} & \multicolumn{2}{c}{\textbf{Sursilvan}} & \multicolumn{2}{c}{\textbf{Sutsilvan}} & \multicolumn{2}{c}{\textbf{Surmiran}} & \multicolumn{2}{c}{\textbf{Puter}} & \multicolumn{2}{c}{\textbf{Vallader}} & \textbf{Avg.} \\
                & from & into & from & into & from & into & from & into & from & into &  \\
\midrule
NLLB-200-1.3B (fine-tuned) & 50.0 & 50.8 & 50.1 & 49.5 & 47.6 & 44.6 & 53.0 & 54.3 & 53.2 & 54.7 & 50.8 \\
\midrule
GPT-4o (few-shot) & 31.0 & 49.3 & 38.0 & 21.8 & 36.8 & 25.5 & 38.7 & 36.9 & 35.2 & 46.1 & 35.9 \\
GPT-4o-mini (few-shot) & 27.5 & 31.8 & 29.5 & 26.2 & 28.7 & 26.0 & 33.3 & 32.9 & 32.9 & 34.9 & 30.4 \\
GPT-4o-mini (fine-tuned) & 35.6 & 39.4 & 37.5 & 32.1 & 34.2 & 34.0 & 41.2 & 41.4 & 41.0 & 42.6 & 37.9 \\
\bottomrule
\end{tabularx}
\caption{Performance of systems translating between Romansh idioms. We report BLEU on a subsample of the \textit{Mediomatix} test split. NLLB-200-1.3B and GPT-4o-mini are fine-tuned with training data from \textit{Mediomatix}. Note that we use different amounts of fine-tuning data: 182,148 sentence pairs for NLLB and 5,000 for GPT-4o-mini.}
\label{tab:mt_results}
\end{table*}

\section{Validation of Alignment Quality}

\subsection{Accuracy on Validation Set}

Table~\ref{tab:valid_f1} reports precision, recall, and F1-score of the alignment on our validation set.
The results show that taking the consensus across all five possible pivot languages yields a higher-precision alignment than using a single pivot language.
Since our goal is to maximize precision, we use the consensus alignment for creating the final parallel corpus.
Detailed validation results for all pivot languages are provided in Appendix~\ref{ap:val_piv_detailed}.



\subsection{Human Evaluation of Precision}
To evaluate the precision of the final aligned corpus, we randomly select 100 rows from the test set of the corpus. The sampled rows contain 472 segments.
We ask a native speaker of Romansh to assess the sample as follows:
\begin{itemize}[topsep=0.4em, itemsep=0em]
    \item If a segment is notably different from the others in the row (e.g., contains less or more information), but is still generally aligned, it should be marked as noise in parallel data.
    \item If a segment is misaligned, it should be marked as an alignment error.
\end{itemize}

\noindent{}Human evaluation shows that 471 out of 472 segments are correctly aligned.
Of the correctly aligned segments, 20 are marked as containing noise.
The noise occurs within 11 rows, meaning that 89\% of the multi-parallel rows are found to be free of noise.
Given that manually evaluated translations in several web-crawled parallel corpora contain 50--83\% correct translations when accounting for each language pair's data size \citep{kreutzer2022quality}, our parallel corpus is relatively high quality. We provide the evaluator instructions in Appendix~\ref{sec:evaluator-instructions}, and examples of the evaluator's qualitative feedback in Appendix~\ref{sec:qual_analysis}.

\section{Machine Translation Experiment}
We demonstrate that the \textit{Mediomatix} corpus can be used to train and evaluate machine translation between Romansh idioms.

\paragraph{Models}
We evaluate a version of NLLB-200-1.3B~\citep{costa-jussa2024nllb} that we fine-tuned on the training split of \textit{Mediomatix}.
In addition, we report results for two commercial LLMs, GPT-4o and GPT-4o-mini~\cite{openai2024gpt4ocard}, in a lower data regime, with either few-shot prompting or fine-tuning on 5000 examples (250 per translation direction).

\paragraph{LLM Prompting}
For prompting GPT-4o and GPT-4o-mini, we use a similar setup as the WMT24 General Machine Translation Shared Task~\cite{kocmi-etal-2024-findings}.\footnote{\url{https://github.com/wmt-conference/wmt-collect-translations}}
We provide the LLMs with 3-shot prompts randomly retrieved from the validation set.
The target idiom is specified in natural language (e.g., \textit{``Translate ... into Sursilvan.''}); see Appendix~\ref{sec:prompt} for the full prompt.
We generate the LLM translations with greedy decoding.

\paragraph{Fine-tuning}
The models were trained in a multilingual fashion, i.e., a single model instance was trained jointly on the 20 translation directions.

For fine-tuning NLLB, we use the full training split of \textit{Mediomatix}, as described in Appendix~\ref{sec:split_counts}.
We add special tokens for each of the five idioms, which we initialize randomly.
Inputs are truncated to 128 tokens.
We train with a peak learning rate of 2e-4 and a batch size of 1,500 for 6 epochs, with early stopping based on validation BLEU.
For decoding we use beam search with size 4.

For fine-tuning GPT-4o-mini, we use default settings recommended by OpenAI (3 epochs, batch size 10, lr multiplier 1.8).

\paragraph{Evaluation Metric}
Due to a lack of support of Romansh in trained metrics such as COMET, we report BLEU~\cite{papineni-etal-2002-bleu}.
Future work could collect human judgments of translation quality and adapt trained metrics to Romansh idioms.

\paragraph{Results}
Table~\ref{tab:mt_results} shows SacreBLEU scores~\cite{papineni-etal-2002-bleu,post-2018-call}\footnote{Signature: \texttt{\#:1|c:mixed|e:no|tok:13a|s:exp|v:2.5}} on a subsample of 500 test examples per translation direction\footnote{\url{https://github.com/ZurichNLP/mediomatix-code/tree/main/mt_experiment/testset_mediomatix}}. BLEU scores are macro-averaged across all source idioms (for ``from'') and target idioms (for ``into'') that are not identical to the given idiom. See Appendix~\ref{ap:detailed_mt} for BLEU scores on the same subset broken down by source- and target-idiom. 
Fine-tuning on a subsample of \textit{Mediomatix} improves the ability of GPT-4o-mini to translate between Romansh idioms, with an average improvement of 7.5 BLEU points over the baseline model, and 2 BLEU points over the larger GPT-4o model.
The performance of an NLLB model fine-tuned on the full training split of \textit{Mediomatix} is much higher (+12.9 BLEU), demonstrating the benefit of the large \textit{Mediomatix} corpus for enabling MT between Romansh idioms.

\section{Conclusion}
The \textit{Mediomatix} corpus is an opportunistic, but relatively large, multi-parallel corpus of text in the five Romansh idioms. For the first time, this corpus allows for the training and evaluation of MT systems that translate between the idioms of the Romansh language. Beyond MT systems for end users, this work will allow for new approaches to data augmentation, expanding the availability of other NLP technology in the idioms of Romansh.

\section*{Limitations}
The corpus creation described in this paper focuses on optimizing precision as opposed to recall, and as a result, a part of the segments in the \textit{Mediomatix} schoolbooks go unused in the aligned corpus (Table~\ref{tab:stats}).
Correspondingly, the human evaluation we perform is limited to an evaluation of precision, to make sure that the segments included in the final corpus are indeed aligned correctly.

A second limitation of this resource is that it currently released under a non-commercial license only.
We are working with the copyright holders to release the corpus under a more permissive license.

The \textit{Mediomatix} schoolbooks for the different idioms were released at different times. The authors of the books report that content in the later additions to the series (i.e., the Surmiran books) is often translated from the earlier additions, meaning the \textit{Mediomatix} dataset contains so-called ``translationese'' to some extent. While this may limit the utility of the corpus for MT \citep{zhang-toral-2019-effect}, other corpora consisting in translated text like \textsc{Flores} have found widespread use \citep{goyal-etal-2022-flores}.


\section*{Author Contributions}
ZH: Data curation, Investigation, Methodology, Software, Writing – original draft, Writing – review \& editing. \\
JV: Conceptualization, Data curation, Funding acquisition, Investigation, Methodology, Project administration, Software, Supervision, Writing – original draft, Writing – review \& editing. \\
AB: Writing – original draft, Writing – review \& editing. \\
AR: Data curation, Investigation, Software. \\
RC: Conceptualization, Resources, Supervision, Validation, Writing – review \& editing. \\
RS: Conceptualization, Funding acquisition, Methodology, Project administration, Supervision, Writing – review \& editing. \\

\section*{Acknowledgements}
ZH, AR and RS acknowledge funding by the Swiss National Science Foundation (project MUTAMUR; no.~213976).
We thank Riccardo Corazza for technical support, and Ignacio Pérez Prat for advice on data licensing.

\bibliography{anthology,custom}

\appendix

\clearpage

\section{Details on Choice of Embedding Model}
\label{sec:val-exp}

We carry out a small-scale alignment experiment with our manually-aligned validation set to determine which model should be used for aligning \textit{Mediomatix}. Specifically, for each of the idiom pairs, we embed the validation set and align a segment in the source idiom to the segment in the target idiom with the highest cosine similarity. We score the alignment in terms of the average proportion of correct one-to-one alignments across all idiom pairs. 

We experiment with the following models for embedding the segments: Qwen3-Embedding-0.6B \citep{qwen3embedding}, sentence-swissBERT \citep{grosjean-vamvas-2024-fine}, OpenAI's text-embedding-3-large\footnote{\url{https://platform.openai.com/docs/guides/embeddings/embedding-models}}, Google's gemini-embedding-exp-03-07 \citep{lee2025gemini}, Voyage AI's voyage-3-large\footnote{\url{https://blog.voyageai.com/2025/01/07/voyage-3-large/}}, and Cohere's embed-v4.0\footnote{\url{https://docs.cohere.com/docs/cohere-embed}}. Besides sentence-swissBERT---for which we explicitly run inference with the Romansh language adapter---we do not find information stating explicitly that Romansh is in the models' pretraining data. As in previous work on compiling parallel corpora for low-resource languages, we expect cross-lingual transfer in the multilingual models to support a reasonable embedding for Romansh data \citep{thompson-koehn-2019-vecalign, schwenk-etal-2021-wikimatrix}. 

As noted in Section \ref{sec:extraction}, we extract the plain text from the textbooks. In the plain text, we retain only the ``<strong>'' markup, as it sometimes provides insight about the meaning of a segment (i.e., in a question and answer with multiple choices, it may distinguish the correct choice). However, we also experiment with embedding the full HTML markup for each segment, as the structural similarities between segments in different textbooks encoded in the HTML may benefit alignment in case the Romansh text embeddings are not of sufficient quality on their own. In addition to embedding each segment's extracted text and full HTML, we experiment with concatenating the HTML and plain text embeddings. Results are in Table \ref{tab:greedy_align}.


\begin{table}[H]
    \centering
    \begin{tabularx}{\linewidth}{@{}Xrrr@{}}
         Model & Text & HTML & Concat \\ \midrule
        cohere-v4 & 96.2 & 94.1 & 95.7 \\
        gemini-embedding & 96.2 & 95.6 & 96.3 \\
        openai-v3 & 93.8 & 90.1 & 93.5 \\
        \mbox{qwen3-Embedding-0.6B} & 94.9 & 94.4 & 95.6 \\
        sentence-swissbert & 73.7 & 70.3 & 77.3 \\
        voyage-v3 & 95.0 & 94.4 & 96.1 \\ \bottomrule
    \end{tabularx}
    \caption{Average proportion of correct alignments across all idiom pairs using different embedding models and text embeddings, HTML embeddings, and their concatenation (`Concat').}
    \label{tab:greedy_align}
\end{table}



\onecolumn

\section{Validation Set Statistics}
\label{ap:val_set_stats}
\begin{table}[H]
\centering
\begin{tabularx}{\textwidth}{@{}Xccccc@{}}
\toprule
 & Segments & Tokens & Single Rows & Deletions & Many Rows \\
\midrule
Sursilvan & 155 & 1855 & 139 & 4 & 8 \\
Sutsilvan & 155 & 1846 & 136 & 6 & 9 \\
Surmiran & 151 & 1952 & 143 & 4 & 4 \\
Puter & 156 & 1849 & 140 & 4 & 7 \\
Vallader & 155 & 1842 & 139 & 4 & 8 \\
\bottomrule
\end{tabularx}
\caption{Descriptive statistics for the manually aligned multi-parallel validation set. ``Single Rows'' refers to alignment rows with just one segment, while ``Many Rows'' refers to alignment rows with more than one segment. ``Deletions'' refers to the number of rows for which the idiom has no parallel segment.}
\label{tab:val_counts}
\end{table}

\section{Multi-parallel Alignment via a Pivot}
\label{ap:mp_fig}
\begin{figure}[h]
    \centering
    \adjustbox{width=0.50\linewidth}{
    \includegraphics[width=\linewidth]{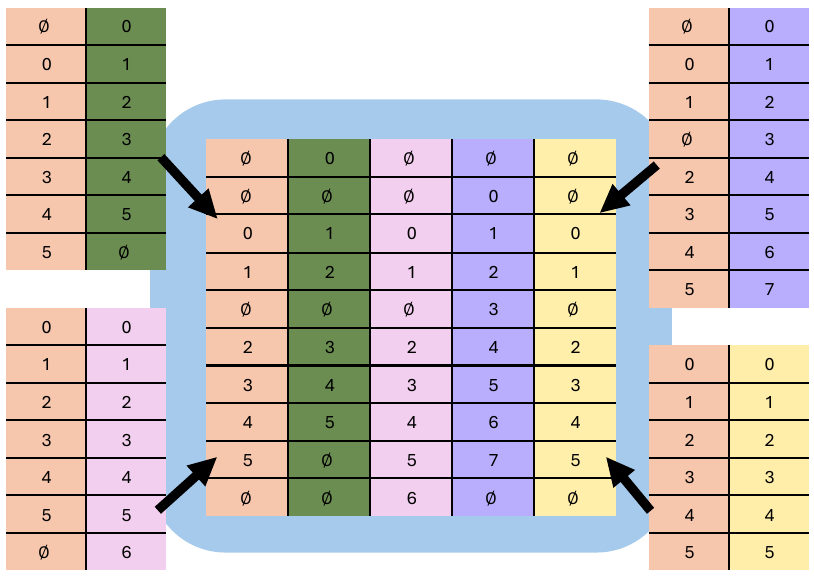}}
    \caption{Depiction of multi-parallel alignment via a pivot idiom, as described in Section \ref{sec:pivot}. The four two-column tables represent bilingual alignments with a given pivot idiom (highlighted in orange). The five-column table shows the result of the full outer join of the bilingual alignments on the pivot idiom, which we take as the multi-parallel alignment for this pivot idiom.}
    \label{fig:pivot_fig}
\end{figure}

\section{Validation Results for Individual Pivots}
\label{ap:val_piv_detailed}

\begin{table*}[h]
    \small
    \centering
    \adjustbox{max width=\textwidth}{
    \begin{tabular}{@{}lcccccc@{}} \toprule
    & \textbf{Sursilvan} & \textbf{Sutsilvan} & \textbf{Surmiran} & \textbf{Puter} & \textbf{Vallader} & \textbf{Consensus} \\ \midrule
cohere-v4 (text) & 94.8/99.2/96.9 & 93.3/98.4/95.7 & 92.3/97.4/94.7 & 95.1/99.6/97.2 & 94.4/99.2/96.7 & \underline{97.2}/\underline{94.1}/\underline{95.4} \\
gemini-embedding (concat) & 93.9/98.6/96.1 & 92.8/98.0/95.2 & 91.2/96.5/93.7 & 95.4/99.8/97.5 & 94.4/99.2/96.7 & 98.1/92.1/94.8 \\
voyage-v3 (concat) & 94.8/99.2/96.9 & 93.7/98.8/96.1 & 93.2/98.2/95.5 & 95.2/99.6/97.2 & 94.4/99.2/96.7 & 97.3/95.6/96.3 \\ \bottomrule
    \end{tabular}}
    \caption{Average strict precision, recall, and F1 values for the validation set pivot alignments using the best three embedding models and input types from the greedy alignment experiment. The column represents the pivot idiom used. \underline{Underlined} scores: Validation set scores for the final configuration used to align \textit{Mediomatix}. We calculate values with the implementation provided with Vecalign, using the \textit{strict} evaluation setup: \url{https://github.com/thompsonb/vecalign}}
    \label{tab:focused_val}
\end{table*}

\section{\textit{Mediomatix} Split Statistics}
\label{sec:split_counts}

\begin{table}[h]
\centering
\begin{tabularx}{\textwidth}{@{}Xrrr@{}} \toprule
\textbf{Idiom} & \textbf{Book Volumes} & \textbf{Aligned Segments} & \textbf{Aligned Tokens} \\ \midrule
Sursilvan & 16 / 10 / 8 / 33 & 12,071 / 7,329 / 4,764 / 25,708 & 103,416 / 80,815 / 57,351 / 271,454 \\
Sutsilvan & 16 / 10 / 8 / 33 & 11,621 / 7,355 / 4,753 / 25,408 & 100,788 / 83,126 / 58,763 / 271,173 \\
Surmiran & 11 / 8 / 8 / 0 & 6,033 / 3,906 / 4,362 / 0 & 54,273 / 38,888 / 52,821 / 0 \\
Puter & 16 / 10 / 8 / 31 & 11,684 / 7,355 / 4,660 / 23,486 & 103,991 / 81,698 / 57,037 / 252,124 \\
Vallader & 16 / 10 / 8 / 31 & 11,654 / 7,391 / 4,680 / 23,672 & 103,039 / 82,583 / 57,214 / 252,568 \\
\bottomrule
\end{tabularx}
\caption{Counts for the train/validation/test/no-rm-surmiran splits of the aligned \textit{Mediomatix} corpus.}
\label{tab:split_counts}
\end{table}

The test and validation splits make up a relatively large portion of the corpus. The primary motivation for splitting along grade levels rather than relegating a certain proportion of segments to each split was to reduce the likelihood of content overlap in the splits. A large validation and test set may also have practical benefits given that in the future, applications of the dataset may include QA or grammar exercise extraction. Large validation and test sets are needed for such tasks to ensure there is sufficient quantity of the exercises in the data used to test systems. 

We also note that in Tables \ref{tab:stats} and \ref{tab:split_counts}, the total number of book volumes in the aligned \textit{Mediomatix} includes fewer volumes relative to the total number available in the schoolbook series (Appendix \ref{sec:book-list}). While manually aligning chapter titles between the books, we observed that some volumes contained no parallel content with the other idioms. In these cases, we dropped those book volumes and only automatically aligned sentences in volumes for idioms that did have largely comparable content.

\clearpage

\section{Qualitative Analysis of \textit{Mediomatix}}
\label{sec:qual_analysis}

\begin{table*}[h]
\begin{tabularx}{\textwidth}{@{}l|XXXX@{}} \toprule
& \textbf{Example 1} & \textbf{Example 2} & \textbf{Example 3} & \textbf{Example 4} \\
\midrule
Sursilvan           & ils 24 da fenadur tochen           & Tgei munta la colur?                    & \textbf{La polizia ei stada tier els a   casa.}            & Jeu mon tuttina a prender penetienzia, schegie che jeu hai fatg ina massa puccaus.  \\ \midrule
Sutsilvan           & igls 24 da fanadur antocen & Tge mùnta la calur?                     & \textbf{La polizeia e stada a controlar igl pass.}       & Jou vont tutegna savens an la stizùn digl mazler, schagea ca jou magl bugent tgarn. \\ \midrule
Surmiran            & \textbf{igls 24 da fanadour}  & Tge monta la calour?                    & \textbf{Las pulicistas òn controllo igl   traffic.}        & \textcolor{red}{Ia magl tschigulatta, perchegl tg'ia va betg gugent ella.}                           \\ \midrule
Puter               & ils 24 lügl fin & \textbf{Che at disch la culur da tias bes-chas?} & La polizia ho controllo il trafic.              & Eau vegn listess a kino, eir scha muossan hoz bgers films.                          \\ \midrule
Vallader            & ils 24 lügl fin                    & \textbf{Che at disch la culur da tias bes-chas?} & La polizia ha controllà il trafic.              & Eu vegn listess a kino, eir schi muossan hoz blers films.                           \\ \midrule
Evaluator Comment & ``\textit{Less Information}''                   & ``\textit{More Information}''                        & ``\textit{Different examples of what the police are doing}'' & ``\textit{Alignment Error}'' \\
\bottomrule
\end{tabularx}
\caption{Example multi-parallel alignments (columns) from the \textit{Mediomatix} corpus. Segments in \textbf{bold} represent segments that were marked as noisy by the evaluator, while the segment in \textcolor{red}{red} text was marked as misaligned by the evaluator.}
\label{tab:eval_examp}
\end{table*}

\noindent{}During their evaluation of the 100 randomly selected multi-parallel rows from \textit{Mediomatix}, the annotator made several remarks. As shown in the second and third column of Table \ref{tab:eval_examp}, they noted that when there was noise in a multi-parallel alignment, the Puter and Vallader segments were usually still parallel. They also noted that in some rows, when segments were not perfectly semantically parallel, they often still demonstrated the same spelling or grammar rule across idioms (i.e., conjugating different verbs for the conditional mood). Feedback such as that in the first two columns of Table \ref{tab:eval_examp} demonstrates a type of alignment noise in which one or more segments in a row contained slightly more or less information than the other segments.

\clearpage

\section{LLM Few-shot Prompt Example}
\label{sec:prompt}

\begin{lstlisting}[
    language=json,
    breaklines=true,
    breakatwhitespace=true,
    columns=fullflexible,
    xleftmargin=2em,
    xrightmargin=1em,
    frame=single
]
[{"role": "user",
  "content": "Translate the following segment surrounded in triple backticks into Vallader. The Sursilvan segment: \n```(*@\textbullet@*) Per tgei va ei en tiu cudisch? Resumescha quei che ti has legiu en treis construcziuns.```\n"},

 {"role": "assistant",
  "content": "```(*@\textbullet@*) Da che tratta teis cudesch? Res(*@\"{u}@*)ma quai cha t(*@\"{u}@*) hast let in trais frasas.```"},

 {"role": "user",
  "content": "Translate the following segment surrounded in triple backticks into Vallader. The Sursilvan segment: \n```4. Jeu hai tschun sutbiadis che stattan el Grischun.```\n"},

 {"role": "assistant",
  "content": "```4. Eu n'ha tschinch bisabiadis chi stan in Grischun.```"},

 {"role": "user",
  "content": "Translate the following segment surrounded in triple backticks into Vallader. The Sursilvan segment: \n```p.ex. alla staziun, alla plazza aviatica, contact cun passagiers jasters```\n"},

 {"role": "assistant",
  "content": "```p.ex. a la staziun, a la plazza aviatica, contact cun passagers esters```"},

 {"role": "user",
  "content": "Translate the following segment surrounded in triple backticks into Vallader. The Sursilvan segment: \n```4. a) Tgeinina ei la differenza denter ils bustabs digl alfabet ed ils suns specials?```\n"}]
\end{lstlisting}

\clearpage

\section{Detailed MT Results}
\label{ap:detailed_mt}

\subsection{NLLB-200-1.3B (fine-tuned)}
\begin{table}[H]
\centering
\begin{tabularx}{\textwidth}{@{}Xrrrrr}
\toprule
\textbf{Source} $\rightarrow$ \textbf{Target} & \textbf{Sursilvan} & \textbf{Sutsilvan} & \textbf{Surmiran} & \textbf{Puter} & \textbf{Vallader} \\
\midrule
\textbf{Sursilvan}   & --   & \cellcolor{uzhblue!34} 59.7   & \cellcolor{uzhblue!20} 44.3   & \cellcolor{uzhblue!23} 48.0   & \cellcolor{uzhblue!23} 48.0 \\
\textbf{Sutsilvan}   & \cellcolor{uzhblue!36} 61.3   & --   & \cellcolor{uzhblue!19} 43.7   & \cellcolor{uzhblue!23} 47.4   & \cellcolor{uzhblue!23} 47.9 \\
\textbf{Surmiran}   & \cellcolor{uzhblue!24} 48.2   & \cellcolor{uzhblue!22} 46.4   & --   & \cellcolor{uzhblue!23} 47.7   & \cellcolor{uzhblue!23} 48.0 \\
\textbf{Puter}   & \cellcolor{uzhblue!22} 46.7   & \cellcolor{uzhblue!21} 45.7   & \cellcolor{uzhblue!21} 44.9   & --   & \cellcolor{uzhblue!48} 75.0 \\
\textbf{Vallader}   & \cellcolor{uzhblue!23} 47.1   & \cellcolor{uzhblue!22} 46.1   & \cellcolor{uzhblue!21} 45.4   & \cellcolor{uzhblue!47} 74.1   & -- \\
\bottomrule
\end{tabularx}
\caption{BLEU scores for machine translation between each pair of Romansh idioms in the \textit{Mediomatix} corpus. Each cell shows the BLEU score for translating from the source idiom (row) to the target idiom (column).}
\label{tab:bleu_scores-nllb}
\end{table}

\subsection{GPT-4o (few-shot)}
\begin{table}[H]
\centering
\begin{tabularx}{\textwidth}{@{}Xrrrrr}
\toprule
\textbf{Source} $\rightarrow$ \textbf{Target} & \textbf{Sursilvan} & \textbf{Sutsilvan} & \textbf{Surmiran} & \textbf{Puter} & \textbf{Vallader} \\
\midrule
\textbf{Sursilvan}   & --   & \cellcolor{uzhblue!2} 24.6   & \cellcolor{uzhblue!3} 25.4   & \cellcolor{uzhblue!10} 33.1   & \cellcolor{uzhblue!17} 40.7 \\
\textbf{Sutsilvan}   & \cellcolor{uzhblue!34} 59.9   & --   & \cellcolor{uzhblue!3} 25.7   & \cellcolor{uzhblue!6} 29.3   & \cellcolor{uzhblue!13} 37.0 \\
\textbf{Surmiran}   & \cellcolor{uzhblue!24} 48.9   & \cellcolor{uzhblue!1} 23.2   & --   & \cellcolor{uzhblue!11} 33.9   & \cellcolor{uzhblue!17} 41.1 \\
\textbf{Puter}   & \cellcolor{uzhblue!20} 43.9   & \cellcolor{uzhblue!0} 19.9   & \cellcolor{uzhblue!3} 25.5   & --   & \cellcolor{uzhblue!40} 65.7 \\
\textbf{Vallader}   & \cellcolor{uzhblue!20} 44.4   & \cellcolor{uzhblue!0} 19.5   & \cellcolor{uzhblue!3} 25.5   & \cellcolor{uzhblue!27} 51.5   & -- \\
\bottomrule
\end{tabularx}
\caption{BLEU scores for machine translation between each pair of Romansh idioms in the \textit{Mediomatix} corpus. Each cell shows the BLEU score for translating from the source idiom (row) to the target idiom (column).}
\label{tab:bleu_scores-gpt4o}
\end{table}

\subsection{GPT-4o-mini (few-shot)}
\begin{table}[H]
\centering
\begin{tabularx}{\textwidth}{@{}Xrrrrr}
\toprule
\textbf{Source} $\rightarrow$ \textbf{Target} & \textbf{Sursilvan} & \textbf{Sutsilvan} & \textbf{Surmiran} & \textbf{Puter} & \textbf{Vallader} \\
\midrule
\textbf{Sursilvan}   & --   & \cellcolor{uzhblue!7} 29.9   & \cellcolor{uzhblue!4} 27.1   & \cellcolor{uzhblue!3} 25.9   & \cellcolor{uzhblue!4} 27.1 \\
\textbf{Sutsilvan}   & \cellcolor{uzhblue!14} 38.0   & --   & \cellcolor{uzhblue!5} 28.0   & \cellcolor{uzhblue!3} 25.4   & \cellcolor{uzhblue!4} 26.7 \\
\textbf{Surmiran}   & \cellcolor{uzhblue!9} 32.3   & \cellcolor{uzhblue!5} 27.4   & --   & \cellcolor{uzhblue!4} 26.2   & \cellcolor{uzhblue!6} 28.7 \\
\textbf{Puter}   & \cellcolor{uzhblue!5} 28.1   & \cellcolor{uzhblue!2} 24.0   & \cellcolor{uzhblue!1} 23.9   & --   & \cellcolor{uzhblue!32} 57.0 \\
\textbf{Vallader}   & \cellcolor{uzhblue!6} 28.9   & \cellcolor{uzhblue!1} 23.6   & \cellcolor{uzhblue!2} 25.0   & \cellcolor{uzhblue!29} 54.1   & -- \\
\bottomrule
\end{tabularx}
\caption{BLEU scores for machine translation between each pair of Romansh idioms in the \textit{Mediomatix} corpus. Each cell shows the BLEU score for translating from the source idiom (row) to the target idiom (column).}
\label{tab:bleu_scores-gpt4o-mini}
\end{table}

\subsection{GPT-4o-mini (fine-tuned)}
\begin{table}[H]
\centering
\begin{tabularx}{\textwidth}{@{}Xrrrrr}
\toprule
\textbf{Source} $\rightarrow$ \textbf{Target} & \textbf{Sursilvan} & \textbf{Sutsilvan} & \textbf{Surmiran} & \textbf{Puter} & \textbf{Vallader} \\
\midrule
\textbf{Sursilvan}   & --   & \cellcolor{uzhblue!16} 39.9   & \cellcolor{uzhblue!12} 35.3   & \cellcolor{uzhblue!10} 32.9   & \cellcolor{uzhblue!11} 34.2 \\
\textbf{Sutsilvan}   & \cellcolor{uzhblue!23} 47.9   & --   & \cellcolor{uzhblue!11} 34.5   & \cellcolor{uzhblue!10} 33.5   & \cellcolor{uzhblue!11} 34.1 \\
\textbf{Surmiran}   & \cellcolor{uzhblue!14} 38.1   & \cellcolor{uzhblue!8} 31.0   & --   & \cellcolor{uzhblue!10} 32.8   & \cellcolor{uzhblue!11} 34.9 \\
\textbf{Puter}   & \cellcolor{uzhblue!12} 35.4   & \cellcolor{uzhblue!6} 29.1   & \cellcolor{uzhblue!10} 33.1   & --   & \cellcolor{uzhblue!41} 67.1 \\
\textbf{Vallader}   & \cellcolor{uzhblue!13} 36.1   & \cellcolor{uzhblue!6} 28.6   & \cellcolor{uzhblue!10} 32.9   & \cellcolor{uzhblue!40} 66.5   & -- \\
\bottomrule
\end{tabularx}

\caption{BLEU scores for machine translation between each pair of Romansh idioms in the \textit{Mediomatix} corpus. Each cell shows the BLEU score for translating from the source idiom (row) to the target idiom (column).}
\label{tab:bleu_scores-gpt4o-mini-ft}
\end{table}

\clearpage

\section{Evaluator Instructions}
\label{sec:evaluator-instructions}
You don't have to edit or proofread the examples. We'd just like to know whether our alignment algorithm worked, i.e., whether the text segments of the different idioms indeed belong together.
\begin{itemize}[topsep=0.4em, itemsep=0em]
\item If the text segments have the same meaning across all five idioms, which is the expected case, do nothing. See Example 1 in the Google Sheet.
\item If the text segments do belong together, but there is an outlier that is highly different in meaning (e.g., contains different information), color it yellow. See Example 2 in the sheet.
\item If the row has an outlier that clearly does not belong with the others, color it red. See Example 3 in the sheet.
\end{itemize}
\noindent{}Some cells will be empty if we did not find a matching segment for an idiom. This is okay and you don't have to mark the empty cells as errors.

\bigskip

\noindent{}\includegraphics[width=\textwidth,trim={0 14cm 4.8cm 0},clip]{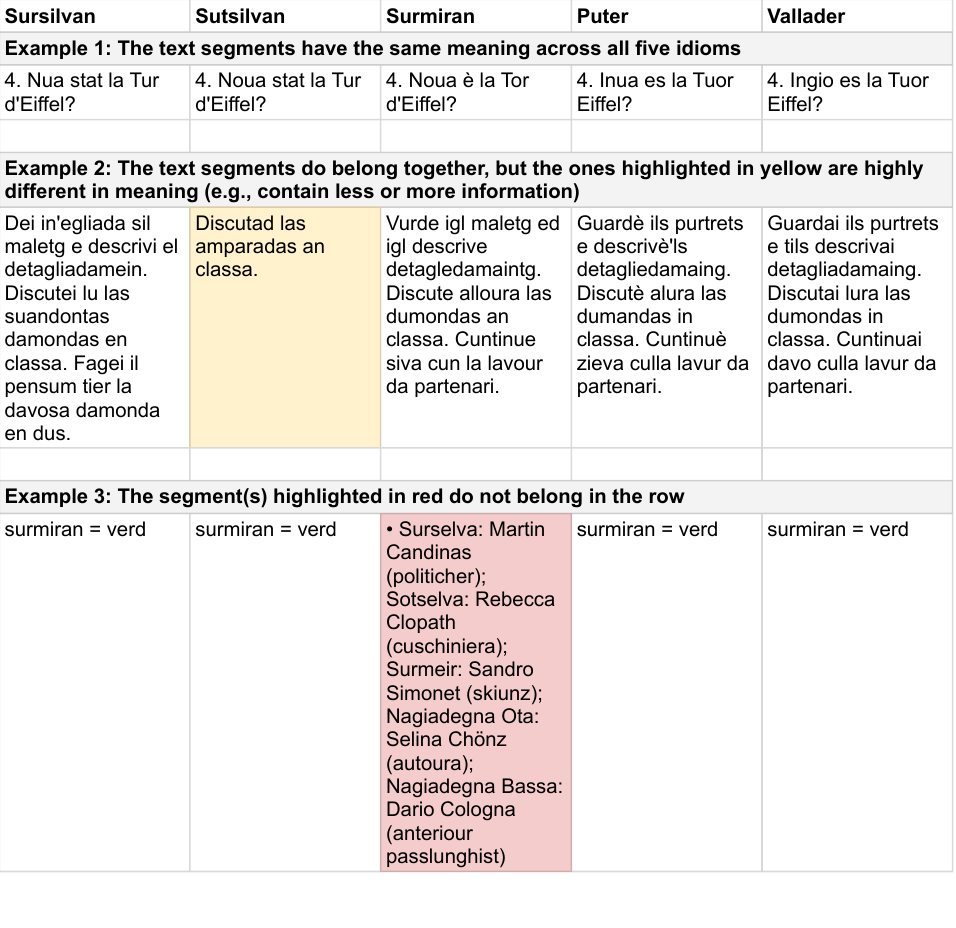}

\clearpage

\section{List of Schoolbooks}
\label{sec:book-list}

Bibliographic data of the schoolbooks included in the corpus, based on \textit{Bündner Bibliografie}.

\noindent{}Each item in the list comprises 4 workbook volumes and 4 separate teacher's commentaries, except schoolbooks for the 4th and 6th grades, which have 5 volumes instead of 4.

{
\small

\subsection*{Sursilvan}
\textbf{ISBN:} 978-3-03847-012-0\\
\textbf{Year:} 2018\\
\textbf{Title:} Mediomatix, 2. classa, lungatg: sursilvan.\\
\textbf{Bibliographic record:} \url{https://www.opac.gr.ch/permalink/41BGR_INST/44cnm/alma990006896790206696}\\[1em]
\textbf{ISBN:} 978-3-03847-016-8\\
\textbf{Year:} 2019\\
\textbf{Title:} Mediomatix, 3. classa, lungatg sursilvan.\\
\textbf{Bibliographic record:} \url{https://www.opac.gr.ch/permalink/41BGR_INST/44cnm/alma990007014990206696}\\[1em]
\textbf{ISBN:} 978-3-03847-020-5\\
\textbf{Year:} 2020\\
\textbf{Title:} Mediomatix, 4. classa, lungatg: sursilvan.\\
\textbf{Bibliographic record:} \url{https://www.opac.gr.ch/permalink/41BGR_INST/44cnm/alma990007140230206696}\\[1em]
\textbf{ISBN:} 978-3-03847-024-3\\
\textbf{Year:} 2021\\
\textbf{Title:} Mediomatix, 5. classa, lungatg: sursilvan.\\
\textbf{Bibliographic record:} \url{https://www.opac.gr.ch/permalink/41BGR_INST/44cnm/alma997476532206696}\\[1em]
\textbf{ISBN:} 978-3-03847-028-1\\
\textbf{Year:} 2021\\
\textbf{Title:} Mediomatix, 6. classa, lungatg: sursilvan.\\
\textbf{Bibliographic record:} \url{https://www.opac.gr.ch/permalink/41BGR_INST/44cnm/alma997476532306696}\\[1em]
\textbf{ISBN:} 978-3-03847-032-8\\
\textbf{Year:} 2020\\
\textbf{Title:} Mediomatix, 1. classa scalem secundar 1, lungatg: sursilvan.\\
\textbf{Bibliographic record:} \url{https://www.opac.gr.ch/permalink/41BGR_INST/44cnm/alma990007112700206696}\\[1em]
\textbf{ISBN:} 978-3-03847-036-6\\
\textbf{Year:} 2019\\
\textbf{Title:} Mediomatix, 2. classa scalem secundar 1, lungatg: sursilvan.\\
\textbf{Bibliographic record:} \url{https://www.opac.gr.ch/permalink/41BGR_INST/44cnm/alma990007015660206696}\\[1em]
\textbf{ISBN:} 978-3-03847-040-3\\
\textbf{Year:} 2018\\
\textbf{Title:} Mediomatix, 3. classa scalem secundar 1, lungatg: sursilvan.\\
\textbf{Bibliographic record:} \url{https://www.opac.gr.ch/permalink/41BGR_INST/44cnm/alma990006897030206696}\\[1em]
\subsection*{Sutsilvan}
\textbf{ISBN:} 978-3-03847-013-7\\
\textbf{Year:} 2018\\
\textbf{Title:} Mediomatix, 2. classa, lungatg: sutsilvan.\\
\textbf{Bibliographic record:} \url{https://www.opac.gr.ch/permalink/41BGR_INST/44cnm/alma990006896810206696}\\[1em]
\textbf{ISBN:} 978-3-03847-017-5\\
\textbf{Year:} 2019\\
\textbf{Title:} Mediomatix, 3. classa, lungatg sutsilvan.\\
\textbf{Bibliographic record:} \url{https://www.opac.gr.ch/permalink/41BGR_INST/44cnm/alma990007015060206696}\\[1em]
\textbf{ISBN:} 978-3-03847-021-2\\
\textbf{Year:} 2020\\
\textbf{Title:} Mediomatix, 4. classa, lungatg: sutsilvan.\\
\textbf{Bibliographic record:} \url{https://www.opac.gr.ch/permalink/41BGR_INST/44cnm/alma990007140200206696}\\[1em]
\textbf{ISBN:} 978-3-03847-025-0\\
\textbf{Year:} 2021\\
\textbf{Title:} Mediomatix, 5. classa, lungatg: sutsilvan.\\
\textbf{Bibliographic record:} \url{https://www.opac.gr.ch/permalink/41BGR_INST/44cnm/alma997476532006696}\\[1em]
\textbf{ISBN:} 978-3-03847-029-8\\
\textbf{Year:} 2021\\
\textbf{Title:} Mediomatix, 6. classa, lungatg: sutsilvan.\\
\textbf{Bibliographic record:} \url{https://www.opac.gr.ch/permalink/41BGR_INST/44cnm/alma997476531806696}\\[1em]
\textbf{ISBN:} 978-3-03847-033-5\\
\textbf{Year:} 2020\\
\textbf{Title:} Mediomatix, 1. classa scalem secundar 1, lungatg: sutsilvan.\\
\textbf{Bibliographic record:} \url{https://www.opac.gr.ch/permalink/41BGR_INST/44cnm/alma990007112660206696}\\[1em]
\textbf{ISBN:} 978-3-03847-037-3\\
\textbf{Year:} 2019\\
\textbf{Title:} Mediomatix, 2. classa scalem secundar 1, lungatg: sutsilvan.\\
\textbf{Bibliographic record:} \url{https://www.opac.gr.ch/permalink/41BGR_INST/44cnm/alma990007015630206696}\\[1em]
\textbf{ISBN:} 978-3-03847-041-0\\
\textbf{Year:} 2018\\
\textbf{Title:} Mediomatix, 3. classa scalem secundar 1, lungatg: sutsilvan.\\
\textbf{Bibliographic record:} \url{https://www.opac.gr.ch/permalink/41BGR_INST/44cnm/alma990006897020206696}\\[1em]
\subsection*{Surmiran}
\textbf{ISBN:} 978-3-03847-122-6\\
\textbf{Year:} 2022\\
\textbf{Title:} Mediomatix. 2, lungatg: surmiran.\\
\textbf{Bibliographic record:} \url{https://www.opac.gr.ch/permalink/41BGR_INST/44cnm/alma997566920606696}\\[1em]
\textbf{ISBN:} 978-3-03847-123-3\\
\textbf{Year:} 2023\\
\textbf{Title:} Mediomatix. 3. classa, lungatg: surmiran.\\
\textbf{Bibliographic record:} \url{https://www.opac.gr.ch/permalink/41BGR_INST/44cnm/alma997922021006696}\\[1em]
\textbf{ISBN:} 978-3-03847-124-0\\
\textbf{Year:} 2024\\
\textbf{Title:} Mediomatix. 4. classa, lungatg: surmiran.\\
\textbf{Bibliographic record:} \url{https://www.opac.gr.ch/permalink/41BGR_INST/44cnm/alma998005220706696}\\[1em]
\textbf{ISBN:} 978-3-03847-125-7\\
\textbf{Year:} 2025\\
\textbf{Title:}  Mediomatix. 5. classa, lungatg: surmiran.\\
\textbf{Bibliographic record:} \url{https://www.opac.gr.ch/permalink/41BGR_INST/44cnm/alma998115845006696}\\[1em]
\subsection*{Puter}
\textbf{ISBN:} 978-3-03847-014-4\\
\textbf{Year:} 2018\\
\textbf{Title:} Mediomatix, 2. classa, lungatg: puter.\\
\textbf{Bibliographic record:} \url{https://www.opac.gr.ch/permalink/41BGR_INST/44cnm/alma990006896830206696}\\[1em]
\textbf{ISBN:} 978-3-03847-018-2\\
\textbf{Year:} 2019\\
\textbf{Title:} Mediomatix, 3. classa, lingua puter.\\
\textbf{Bibliographic record:} \url{https://www.opac.gr.ch/permalink/41BGR_INST/44cnm/alma990007015160206696}\\[1em]
\textbf{ISBN:} 978-3-03847-022-9\\
\textbf{Year:} 2020\\
\textbf{Title:} Mediomatix, 4. classa, lingua: puter.\\
\textbf{Bibliographic record:} \url{https://www.opac.gr.ch/permalink/41BGR_INST/44cnm/alma990007140210206696}\\[1em]
\textbf{ISBN:} 978-3-03847-026-7\\
\textbf{Year:} 2021\\
\textbf{Title:} Mediomatix, 5. classa, lingua: puter.\\
\textbf{Bibliographic record:} \url{https://www.opac.gr.ch/permalink/41BGR_INST/44cnm/alma997476531706696}\\[1em]
\textbf{ISBN:} 978-3-03847-030-4\\
\textbf{Year:} 2021\\
\textbf{Title:} Mediomatix. 6. classa lingua: puter.\\
\textbf{Bibliographic record:} \url{https://www.opac.gr.ch/permalink/41BGR_INST/44cnm/alma997476531606696}\\[1em]
\textbf{ISBN:} 978-3-03847-034-2\\
\textbf{Year:} 2020\\
\textbf{Title:} Mediomatix, 1. classa s-chelin secundar 1, lingua: puter.\\
\textbf{Bibliographic record:} \url{https://www.opac.gr.ch/permalink/41BGR_INST/44cnm/alma990007112690206696}\\[1em]
\textbf{ISBN:} 978-3-03847-038-0\\
\textbf{Year:} 2019\\
\textbf{Title:} Mediomatix, 2. classa s-chelin secundar 1, lungatg: puter.\\
\textbf{Bibliographic record:} \url{https://www.opac.gr.ch/permalink/41BGR_INST/44cnm/alma990007015590206696}\\[1em]
\textbf{ISBN:} 978-3-03847-042-7\\
\textbf{Year:} 2018\\
\textbf{Title:} Mediomatix, 3. classa s-chelin secundar 1, lungatg: puter.\\
\textbf{Bibliographic record:} \url{https://www.opac.gr.ch/permalink/41BGR_INST/44cnm/alma990006896870206696}\\[1em]
\subsection*{Vallader}
\textbf{ISBN:} 978-3-03847-015-1\\
\textbf{Year:} 2018\\
\textbf{Title:} Mediomatix, 2. classa, lungatg: vallader.\\
\textbf{Bibliographic record:} \url{https://www.opac.gr.ch/permalink/41BGR_INST/44cnm/alma990006896860206696}\\[1em]
\textbf{ISBN:} 978-3-03847-019-9\\
\textbf{Year:} 2019\\
\textbf{Title:} Mediomatix, 3. classa, lingua vallader.\\
\textbf{Bibliographic record:} \url{https://www.opac.gr.ch/permalink/41BGR_INST/44cnm/alma990007014880206696}\\[1em]
\textbf{ISBN:} 978-3-03847-023-6\\
\textbf{Year:} 2020\\
\textbf{Title:} Mediomatix, 4. classa, lingua: vallader.\\
\textbf{Bibliographic record:} \url{https://www.opac.gr.ch/permalink/41BGR_INST/44cnm/alma990007140190206696}\\[1em]
\textbf{ISBN:} 978-3-03847-027-4\\
\textbf{Year:} 2021\\
\textbf{Title:} Mediomatix, 5. classa, lingua: vallader.\\
\textbf{Bibliographic record:} \url{https://www.opac.gr.ch/permalink/41BGR_INST/44cnm/alma997481737406696}\\[1em]
\textbf{ISBN:} 978-3-03847-031-1\\
\textbf{Year:} 2021\\
\textbf{Title:} Mediomatix, 6. classa, lingua: vallader.\\
\textbf{Bibliographic record:} \url{https://www.opac.gr.ch/permalink/41BGR_INST/44cnm/alma997481737306696}\\[1em]
\textbf{ISBN:} 978-3-03847-035-9\\
\textbf{Year:} 2020\\
\textbf{Title:} Mediomatix, 1. classa s-chalin secundar 1, lingua: vallader.\\
\textbf{Bibliographic record:} \url{https://www.opac.gr.ch/permalink/41BGR_INST/44cnm/alma990007112650206696}\\[1em]
\textbf{ISBN:} 978-3-03847-039-7\\
\textbf{Year:} 2019\\
\textbf{Title:} Mediomatix, 2. classa s-chalin secundar 1, lingua: vallader.\\
\textbf{Bibliographic record:} \url{https://www.opac.gr.ch/permalink/41BGR_INST/44cnm/alma990007015710206696}\\[1em]
\textbf{ISBN:} 978-3-03847-043-4\\
\textbf{Year:} 2018\\
\textbf{Title:} Mediomatix, 3. classa s-chalin secundar 1, lungatg: vallader.\\
\textbf{Bibliographic record:} \url{https://www.opac.gr.ch/permalink/41BGR_INST/44cnm/alma990006897040206696}
}

\vfill

\end{document}